\documentclass[letterpaper]{article} % DO NOT CHANGE THIS
\usepackage{aaai2026}  % DO NOT CHANGE THIS
\usepackage{times}  % DO NOT CHANGE THIS
\usepackage{helvet}  % DO NOT CHANGE THIS
\usepackage{courier}  % DO NOT CHANGE THIS
\usepackage[hyphens]{url}  % DO NOT CHANGE THIS
\usepackage{graphicx} % DO NOT CHANGE THIS
\usepackage{natbib}  % DO NOT CHANGE THIS AND DO NOT ADD ANY OPTIONS TO IT
\usepackage{caption} % DO NOT CHANGE THIS AND DO NOT ADD ANY OPTIONS TO IT
\usepackage{algorithm}
\usepackage{algorithmic}
\usepackage{amsmath}
\usepackage{booktabs}
\usepackage{amsfonts}
\usepackage{multirow}
\usepackage{booktabs} %
\usepackage{tikz}
\usepackage{pgfplots}
\usetikzlibrary{positioning}
\usetikzlibrary{calc}
\usepackage{subcaption}
\usepackage{etoolbox,lipsum}
\usepackage{xcolor}
\usepackage{afterpage}%\usepackage{wrapfig}
\usepackage{graphicx}
\graphicspath{{./}{LaTeX/}}

\usepackage{newfloat}
\usepackage{listings}
\DeclareCaptionStyle{ruled}{labelfont=normalfont,labelsep=colon,strut=off} % DO NOT CHANGE THIS
\floatstyle{ruled}
\newfloat{listing}{tb}{lst}{}
\floatname{listing}{Listing}
\title{RMFAT: Recurrent Multi-scale Feature Atmospheric Turbulence Mitigator}
\author{
    Zhiming Liu and Nantheera Anantrasirichai
}
\affiliations{
Visual Information Laboratory\\
University of Bristol, Bristol, UK
}

\usepackage{bibentry}
\begin{document}

\maketitle

\begin{abstract}
Atmospheric turbulence severely degrades video quality by introducing distortions such as geometric warping, blur, and temporal flickering, posing significant challenges to both visual clarity and temporal consistency. Current state-of-the-art methods are based on transformer, 3D architectures and require multi-frame input, but their large computational cost and memory usage limit real-time deployment, especially in resource-constrained scenarios. In this work, we propose RMFAT — Recurrent Multi-scale Feature Atmospheric Turbulence Mitigator designed for efficient and temporally consistent video restoration under AT conditions. RMFAT adopts a lightweight recurrent framework that restores each frame using only two inputs at a time, significantly reducing temporal window size and computational burden. It further integrates multi-scale feature encoding and decoding with temporal warping modules at both encoder and decoder stages to enhance spatial detail and temporal coherence. Extensive experiments conducted on synthetic and real-world atmospheric turbulence datasets demonstrate that RMFAT not only outperforms existing methods in terms of clarity restoration (with nearly a 9\% improvement in SSIM) but also achieves significantly improved inference speed (achieving a more than fourfold reduction), making it particularly suitable for real-time atmospheric turbulence suppression tasks.
\end{abstract}

\section{Introduction}

Atmospheric turbulence (AT), caused by temperature gradients in the atmosphere, is a common phenomenon that severely degrades the quality of captured videos. It introduces spatiotemporal distortions such as geometric warping, blurring, and flickering, which not only reduce visual fidelity but also hinder downstream tasks like object detection and recognition. Traditional image processing methods, including deconvolution~\cite{Deledalle:blind:2020} and lucky imaging~\cite{6471221}, often fail in dynamic environments due to their reliance on assumptions such as global motion or static scenes.

\begin{figure}[t!]
    \centering
    \includegraphics[width=0.95\linewidth]{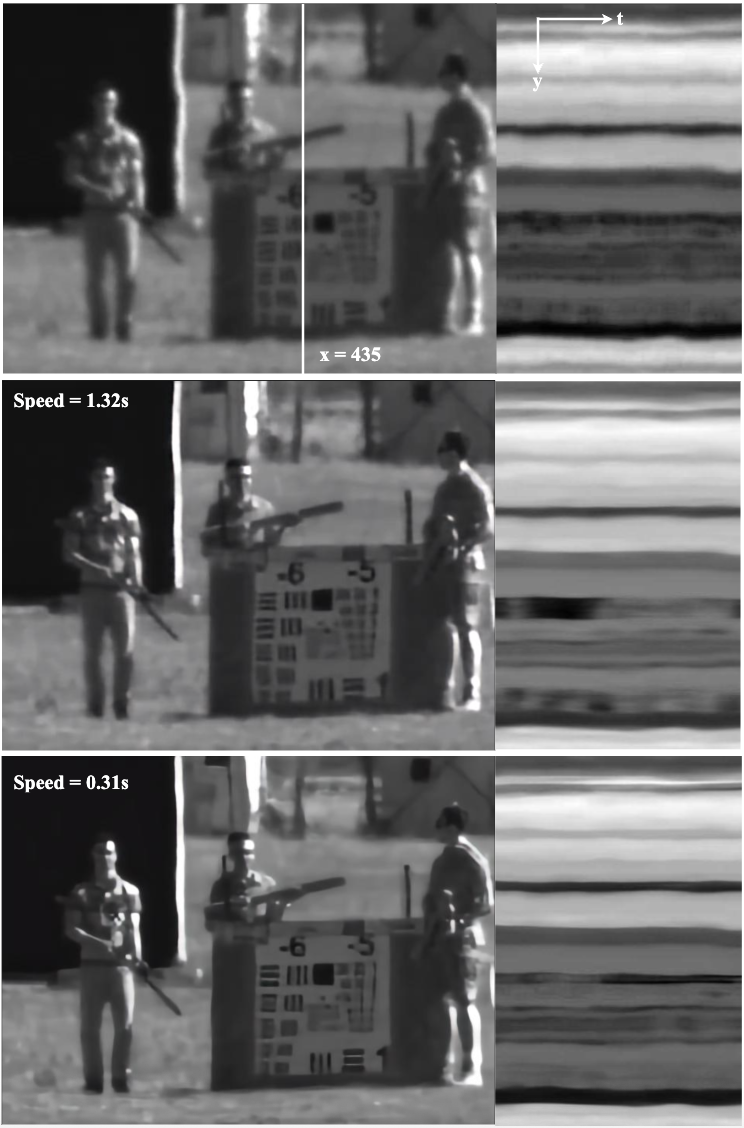}
    \caption{
    Results of the \textit{People with tools} sequence. Top to bottom: distorted input, multi-frame restoration results, and our recurrent technique, which also achieves temporal smoothing but is significantly faster. The left column shows Frame $t = 45$, while the right column displays $y$--$t$ slices at $x = 435$ (first 90 frames).
    }
    \label{fig:feature}
\end{figure}

\begin{figure*}[t!]
    \centering
    \includegraphics[width=1.0\linewidth]{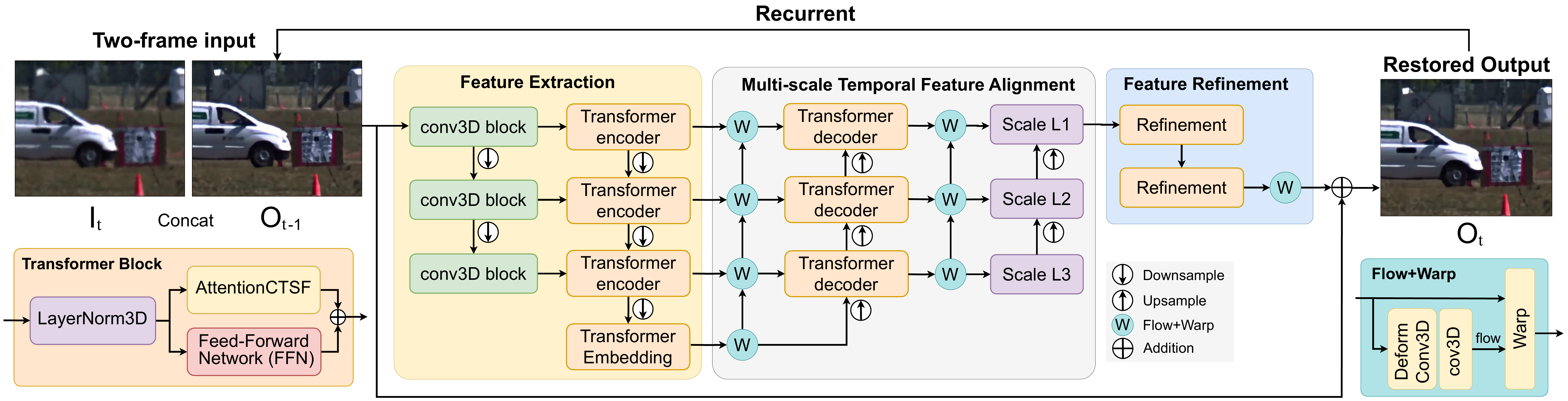}
    \caption{Overview of our proposed recurrent turbulence restoration framework. Given the current degraded frame $I_t$ and the previously restored output $O_{t-1}$, the model first performs feature extraction using 3D convolutions and Transformer encoders at multiple scales. These features are temporally aligned through warping and processed by Transformer decoders. The aligned multi-scale features are then fused and refined through a dedicated refinement module. The Transformer block structure (bottom-left) includes LayerNorm3D, self-attention (AttentionCTSF), and a feed-forward network. The model is recurrent across time, where each restored output $O_t$ is fed into the next step ($t+1$). The bottom-right illustrates the flow and warp module, which leverages deformable 3D convolutions.
    }
    \label{fig:recurrent_pipeline}
\end{figure*}

Recent advances in deep learning have led to significant improvements in atmospheric turbulence mitigation. Transformer-based models~\cite{zhang2023imagingatmosphereusingturbulence,Zhicheng:deturb:2024,Hill2025}. These methods typically suffer from high computational cost and slow inference speeds, making them impractical for real-time or resource-constrained applications. Mamba-based method~\cite{hill2025mamat3dmambabasedatmospheric} 
has shown strong performance in capturing long-range dependencies and reduced complexity. However, it remains limited for real-time use due to its reliance on large multi-frame input. Moreover, since pixels are displaced from their original positions due to atmospheric turbulence typically with approximately zero mean and quasi-periodic motion \cite{li2009suppressing, guo2024direct}, using more multiple frames can help reduce distortion more effectively. However, limited memory capacity constrains the number of frames that can be processed, which in turn limits the achievable restoration quality.

In response to these challenges, we propose \textbf{R}ecurrent \textbf{M}ulti-scale \textbf{F}eature \textbf{A}tmospheric \textbf{T}urbulence Mitigator (\textbf{RMFAT}), a lightweight and efficient video restoration model designed for temporally consistent atmospheric turbulence mitigation. RMFAT adopts a recurrent inference mechanism that progressively restores distorted frames using only the current distorted input  and the previously restored output. This recurrent design enables improved restoration by effectively aggregating information from earlier frames, akin to applying exponentially decaying weights to past outputs. Fig.~\ref{fig:feature} demonstrates the effectiveness of the recurrent approach in both sharpness and temporal smoothing, compared to the multi-frame input method (using 10 frames, limited by the maximum memory of the RTX 4090). The recurrent inference framework achieves high-quality restoration with significantly lower memory usage and computational demands, enabling real-time deployment in applications where low latency and high temporal stability are essential, such as remote sensing and surveillance scenarios.

Moreover, by integrating multi-scale feature encoding and decoding, RMFAT effectively captures and fuses spatiotemporal information at multiple levels of abstraction. To further enhance temporal coherence, we incorporate optical flow-based feature warping at both encoder and decoder stages, enabling precise alignment across frame. These flow fields are also accumulated to generate a series of warped past outputs, which are then used to compute a novel temporal consistency loss. This loss encourages the aggregation of restored features from previous frames while simultaneously enforcing temporal smoothness.

%Unlike previous works that require multiple frames and heavy architectures for restoration, RMFAT operates efficiently in a recurrent manner with minimal frame input and computational overhead. 

In summary, our key contributions are as follows:
\begin{itemize}
\item We propose RMFAT, a novel lightweight and efficient video restoration model for mitigating atmospheric turbulence.
\item Our recurrent inference framework integrates temporal alignment and warping into a multi-scale transformer architecture, enabling it to handle dynamic scenes with varying levels of turbulence while supporting low-latency, memory-efficient processing suitable for real-time applications.
\item We introduce a novel temporal consistency loss that encourages smooth and coherent restoration over time by leveraging accumulated flow-warped past outputs.
\end{itemize}

Experimental results on both synthetic and real-world turbulent video datasets demonstrate that RMFAT achieves state-of-the-art restoration quality while being significantly faster than prior methods. The combination of recurrent inference, multi-scale transformer blocks, and temporal alignment enables RMFAT to restore temporally coherent and visually clear videos even under severe turbulence.

\section{Related Work}

%\paragraph{Atmospheric turbulence mitigation.}
Over the past decade, deep learning has driven the development of numerous turbulence mitigation methods. Early approaches were primarily based on CNN architectures~\cite{gao2019atmospheric, mao2021acceleratingatmosphericturbulencesimulation, anan2023atmospheric}, but recent work has demonstrated the superior capability of transformers in capturing complex distortion patterns. TurbNet~\cite{mao2022single} adopts a UNet-like architecture, replacing convolutional layers with transformer blocks. TMT~\cite{zhang2023imagingatmosphereusingturbulence} introduces a two-stage pipeline—de-tilting to correct local shifts and de-blurring to enhance clarity—guided by a multi-scale loss to improve performance under varying turbulence levels. Wo et al.~\cite{Wu:24} enhance local-global context interaction using sliding window-based self-attention and channel attention, inspired by phase distortion and point spread function representations. 

Physics-inspired models have also gained traction for turbulence removal~\cite{Jaiswal:Physics:2023, Jiang:NeRT:2023, peters2025structured}. Diffusion models have shown superior performance on single-image restoration~\cite{Nair:AT-DDPM:2023}, while transformer-based methods remain state-of-the-art for video restoration~\cite{Anantrasirichai:Artificial:2021}. The DATUM framework~\cite{zhang2024spatiotemporalturbulencemitigationtranslational} replicates classical mitigation pipelines~\cite{6471221}, combining deformable attention for frame alignment with recursive features and residual dense blocks for fusion. \citet{Saha_2024_CVPR} separates moving objects from the static background and applies an image restoration transformer model trained on synthetic data with tilt and adaptive blur.

Recently, MambaTM~\cite{zhang2025phase} and MAMAT~\cite{hill2025mamat3dmambabasedatmospheric} integrates 3D Mamba within the multi-scale encoders, outperforming both TMT and DATUM on synthetic benchmarks. However, these models typically rely on multi-frame input, large-scale computation, and heavy architectural design, limiting their applicability in real-time or resource-constrained scenarios. In contrast to these heavy designs, our RMFAT adopts a lightweight recurrent transformer framework that restores frames using only two inputs, while still achieving competitive performance via multi-scale fusion and temporal alignment.


\section{Proposed Method}
We propose RMFAT, a lightweight Transformer-based recurrent video restoration framework to address image degradation caused by atmospheric turbulence. The overall architecture is illustrated in Fig.~\ref{fig:feature} and consists of three key components: feature extraction, multi-scale temporal feature alignment, and a feature refinement module.
As discussed in the previous section, existing methods typically rely on  multi-frame inputs to enhance restoration quality. However, this often leads to high memory consumption and training complexity limit its applicability. To address this, we reformulate the multi-frame restoration task as a recurrent two-frame restoration problem. This approach significantly reduces computational overhead while enhancing both restoration quality and temporal consistency, making the model more efficient and scalable.

\subsection{Two-Frame Recurrent Restoration}
Our recurrent mechanism explicitly concatenates the current degraded frame with the previously restored frame along the temporal dimension. This combined input enables the model to learn temporal dependencies efficiently.
Specifically, at each time step $t$, the model receives the current degraded frame $I_t$ and the previous restoration result $\hat{O}_{t-1}$, which are concatenated to form a 5D tensor $X_t \in \mathbb{R}^{B \times C \times 2 \times H \times W}$. This tensor serves as the model input for predicting the current restored frame $\hat{O}_t$. This process is performed recurrently, restoring the entire video frame by frame. By incorporating the restoration result of the previous frame into the current prediction, the model can transmit and accumulate information over time, thereby gradually improving its understanding of scene dynamics. Unlike traditional Transformer models that rely on long sequence inputs, our two-frame recurrent structure significantly reduces memory and computational resource consumption while maintaining strong temporal modelling capabilities.

\subsection{Enhanced Temporal Multi-Scale Transformer}

The overall network adopts a hierarchical encoding-decoding structure, comprising multi-scale feature extraction, temporal feature alignment, and fine-grained refinement modules, with the specific structure shown in Figure~\ref{fig:recurrent_pipeline}.

During the feature extraction stage, three layers of 3D convolutional modules downsample the input sequence to extract multi-scale spatio-temporal features. These features are then fed into a Transformer encoder to further refine the spatio-temporal context representation and are fused into a unified feature space via Transformer Embedding.

To enhance inter-frame consistency, we introduce a multi-scale temporal feature alignment module. In contrast to existing approaches, e.g. \citet{zhang2023imagingatmosphereusingturbulence, Zhicheng:deturb:2024}, which apply alignment mechanisms only within the encoder, our method performs non-rigid registration (Flow-Warp modules in Fig. \ref{fig:recurrent_pipeline}) at all three  decoding scales (L1, L2, L3). This design enables dynamic alignment of historical frames throughout the decoder pathway. The aligned features are then decoded and fused with the backbone network, significantly improving temporal coherence across frames and effectively mitigating temporal artifacts during reconstruction.

After decoding, we further refine the output using two Transformer-based refinement modules. Each module comprises a LayerNorm3D layer, an AttentionCTSF multi-head attention mechanism, and a feed-forward network, enabling fine-grained enhancement of features to suppress residual noise. A residual connection is employed, adding the refined output to the original degraded input frame. This design allows the model to concentrate on restoring high-frequency details while avoiding redundant modeling of low-frequency content already present in the input, thereby enhancing both training stability and restoration quality.

\subsection{Loss Functions}

Our training objective integrates multiple loss terms designed to jointly enhance reconstruction fidelity, frequency alignment, temporal consistency, and semantic preservation. The total loss is formulated as:
\begin{equation}
\mathcal{L}_{\text{total}} = \mathcal{L}_{\text{charb}} + \lambda_{\text{dwt}} \mathcal{L}_{\text{DWT}} + \lambda_{\text{flow}} \mathcal{L}_{\text{flow}} + \lambda_{\text{det}} \mathcal{L}_{\text{det}},
\end{equation}
where $\lambda_{\text{dwt}}, \lambda_{\text{flow}}, \lambda_{\text{det}}$ are weighting coefficients.

\subsubsection{Charbonnier Loss}

To supervise pixel-level restoration quality, we use the Charbonnier loss between the restored frame $\hat{O}_t$ and the corresponding ground truth $T_t$. The Charbonnier loss is a differentiable variant of the $\ell_1$ loss with improved robustness to outliers, defined as:
\begin{equation}
\mathcal{L}_{\text{charb}} = \sqrt{||\hat{O}_t - T_t||^2 + \epsilon^2},
\end{equation}
where $\epsilon$ is a small constant (typically $10^{-3}$) to ensure numerical stability.

\subsubsection{Wavelet Loss}

To enforce structural consistency in the frequency domain, we compute a Discrete Wavelet Transform (DWT) loss. Both the restored frame and ground truth are decomposed into multi-scale frequency sub-bands via wavelet transforms. The loss is computed as the Charbonnier distance between corresponding sub-bands:
\begin{equation}
\mathcal{L}_{\text{DWT}} = \sum_{i=1}^{L} \sqrt{||W_i(\hat{O}_t) - W_i(T_t)||^2 + \epsilon^2},
\end{equation}
where $W_i(\cdot)$ denotes the $i$-th wavelet sub-band and $L$ is the total number of sub-bands. This term guides the model to recover high-frequency details and texture.

\subsubsection{Flow-based Loss}

To enforce smooth temporal transitions, we introduce a flow-based temporal consistency loss. For each training step, the model retains the past $K$ restored outputs $\{\hat{O}_{t-k}\}_{k=1}^{K}$, and computes optical flow fields $F_{t-k \rightarrow t}$ between $\hat{O}_{t-k}$ and $\hat{O}_t$. These flows are obtained by accumulating optical flows from the Flow-Warp modules. The past outputs are then warped to the current frame using these flows:
\begin{equation}
\hat{O}^{(k)}_{t} = \text{Warp}(\hat{O}_{t-k}, F_{t-k \rightarrow t}),
\end{equation}
and compared with $\hat{O}_t$ via an $\ell_1$ distance with exponentially decaying weights:
\begin{equation}
\mathcal{L}_{\text{flow}} = \sum_{k=1}^{K} \lambda_k \cdot \| \hat{O}_t - \hat{O}^{(k)}_{t} \|_1, \quad \lambda_k = 0.5^k.
\end{equation}
This penalizes inconsistent restoration across time and promotes frame-to-frame coherence.

\subsubsection{Detection Loss}
To ensure the restoration model preserves semantically meaningful regions, particularly for human-centric scenes, we incorporate a detection-guided auxiliary loss using a pretrained YOLO detector. On the final frame $\hat{O}_T$ of each input sequence, we apply YOLO to predict object bounding boxes $\{B_i\}$ and confidence scores $\{c_i\}$. These predictions are supervised using ground truth boxes $\{G_j\}$ for the "person" class. The detection loss includes two components:

\textbf{(i) IoU Regression Loss:} We compute a soft IoU loss for matched predicted and ground-truth boxes:
\begin{equation}
\mathcal{L}_{\text{IoU}} = 1 - \text{IoU}(B_i, G_j),
\end{equation}

\textbf{(ii) Confidence Loss:} A binary cross-entropy loss is applied to the objectness confidence score:
\begin{equation}
\mathcal{L}_{\text{conf}} = - [y \cdot \log(c_i) + (1 - y) \cdot \log(1 - c_i)],
\end{equation}
where $y = 1$ if a person is present in the ground truth.

The final detection-aware loss is:
\begin{equation}
\mathcal{L}_{\text{det}} = \mathcal{L}_{\text{IoU}} + \mathcal{L}_{\text{conf}}.
\end{equation}

This term encourages the restoration network to preserve object-level integrity, especially for people under severe degradation.

\section{Experiments And Results}

\subsection{Dataset and experiment settings}
\subsubsection{Synthetic datasets.} To evaluate the proposed method under diverse conditions, we construct two synthetic datasets named Static and Dynamic, based on the COCO2017 dataset~\cite{lin2015microsoftcococommonobjects} and GOT-10k dataset~\cite{huang2021got10k}, respectively. The Phase-to-Space (P2S) transform~\cite{mao2021acceleratingatmosphericturbulencesimulation} is employed to simulate realistic atmospheric turbulence, including blur, geometric distortion, and intensity fluctuation.
In the Static setting, each COCO image is converted into a 50-frame turbulence sequence with a fixed background. In the Dynamic setting, the P2S transform is directly applied to dynamic  GOT-10k video clips to simulate turbulence in motion scenarios.
A summary of the dataset specifications is provided in Table~\ref{tab:dataset}.

\begin{table}[t]
\centering
\begin{tabular}{lcc}
\toprule
 & \textbf{Static} & \textbf{Dynamic} \\
\midrule
Source & COCO2017 & GOT-10k \\
%Amount & 5,506 videos & 2,204 videos \\
Training & 5,002 videos & 1,762 videos \\
Validation & 504 videos & 442 videos \\
\bottomrule
\end{tabular}
\caption{Specification of the synthetic turbulence datasets constructed using COCO2017 and GOT-10k.}
\label{tab:dataset}
\end{table}

\subsubsection{Experiment settings.}
We train our restoration model using a two-frame recurrent approach over 100 epochs. Input patches are cropped to $256 \times 256$ with a batch size of 1, and training is conducted on synthetic turbulent sequences. The optimizer is Adam with an initial learning rate of $1 \times 10^{-4}$, adjusted by a StepLR scheduler that halves the rate every 5 epochs to encourage stable convergence.
%Our total loss is composed of multiple components: (1) the Charbonnier loss for pixel-wise reconstruction, (2) a discrete wavelet transform (DWT) loss to enforce frequency consistency, (3) a flow-based temporal consistency loss computed by warping previous outputs using estimated optical flows across frames, and (4) an auxiliary person detection loss using a pretrained YOLOv11n detector applied to the final frame of each sequence. The detection loss is computed only once per sequence to avoid unnecessary redundancy. 
We apply a dynamic weighting scheme to the temporal loss and detection loss, increasing their influence as training progresses. For validation, we report PSNR and SSIM metrics averaged over all frames in the evaluation set.

To further enhance temporal robustness, a buffer of past outputs is maintained and recursively aligned with the current output using multi-scale optical flow estimation. Models are saved based on the best combined PSNR and SSIM score observed during validation.

To comprehensively evaluate the effectiveness and efficiency of our proposed RMFAT model, we conducted comparative experiments against state-of-the-art methods on both synthetic and real datasets, using both objective and subjective assessments. The compared methods include: BasicVSR++~\cite {Chan_2022_CVPR}, TSRWGAN~\cite{jin_chen:Neutralizing:2021}, TMT~\cite{10400926}, DATUM~\cite{Zhang_2024_CVPR},  DeTurb~\cite{Zhicheng:deturb:2024}, MAMAT~\cite{hill2025mamat3dmambabasedatmospheric} and MambaTM~\cite{Zhang_2025_CVPR}.

\begin{figure}
    \centering
    \includegraphics[width=\linewidth]{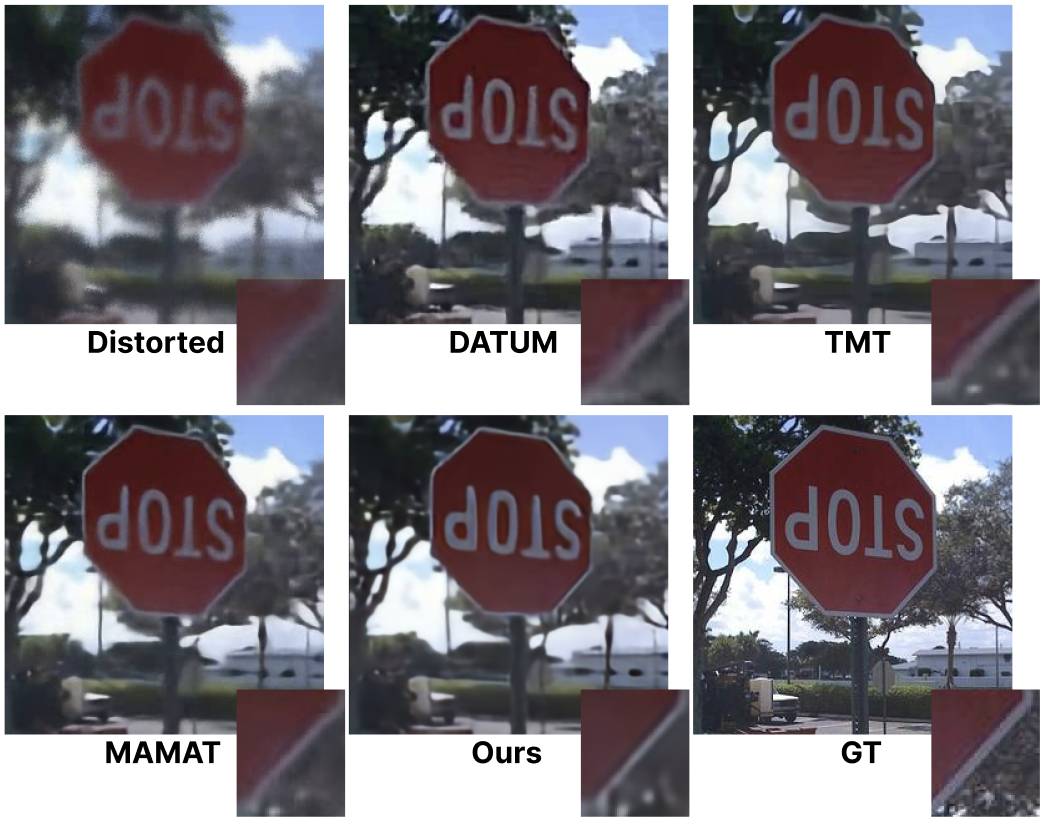}
    \caption{Subjective comparison on a synthetic turbulence-distorted frame.}
    \label{fig:synthetic}
\end{figure}
\begin{figure*}[h]
    \centering
    \includegraphics[width=1.0\linewidth]{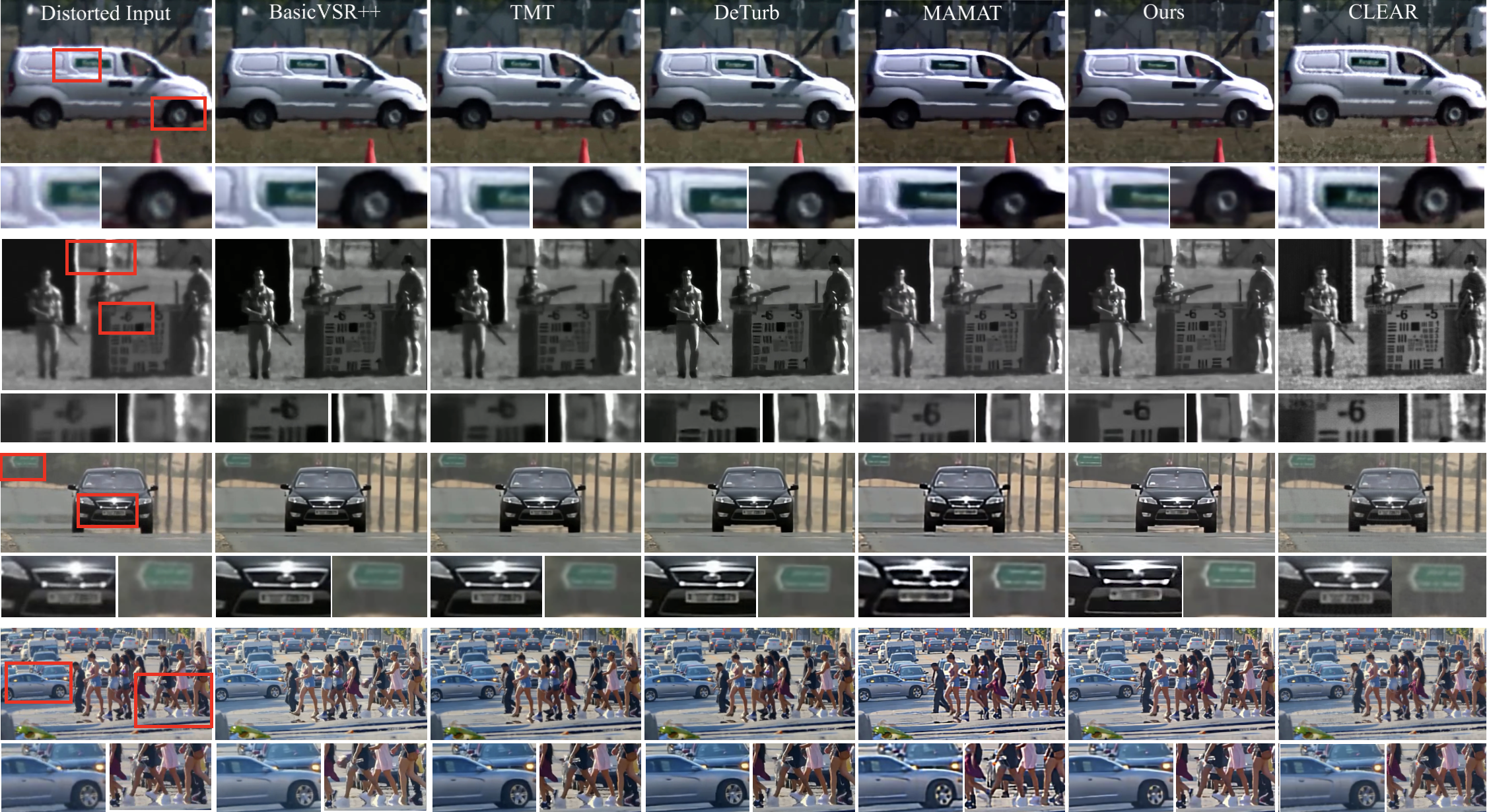}
    \caption{Qualitative comparisons on real-world turbulence-distorted frames from the \textbf{CLEAR} dataset. The results are shown in the following order: raw input, prior methods (BasicVSR++, TMT, DeTurb, MAMAT), our RMFAT, and ground truth (CLEAR)}.
    \label{fig:real}

    \centering
    \includegraphics[width=1.0\linewidth]{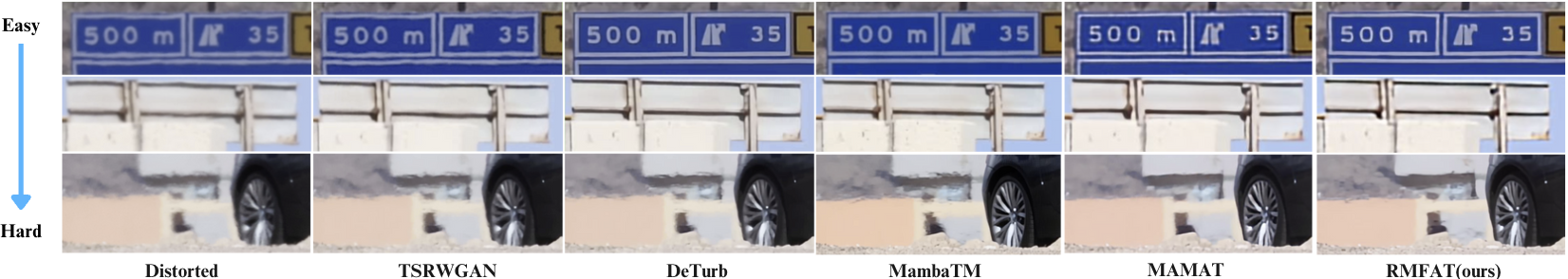}
    \caption{Qualitative comparison on real-world \textbf{ATD} scenes with increasing turbulence severity (top to bottom). From left to right: raw input, prior methods (TSRWGAN, DeTurb, MambaTM, MAMAT), and our RMFAT.}
    \label{fig:easytohard}

\end{figure*}
\begin{table*}[t]
\setlength{\tabcolsep}{2pt}
\centering
\resizebox{\textwidth}{!}{
\begin{tabular}{@{}l|cc|ccc|ccc|ccc|ccc@{}}
\toprule
\multirow{3}{*}{\textbf{Method}} & \multirow{3}{*}{\shortstack{\textbf{Params} \\ \textbf{(M)}~$\downarrow$}}& \multirow{3}{*}{\shortstack{\textbf{Runtime} \\ \textbf{(s)}~$\downarrow$}} 
& \multicolumn{6}{c|}{\textbf{Synthetic Dataset}} 
& \multicolumn{6}{c}{\textbf{Real Dataset}} \\
\cmidrule(lr){4-9} \cmidrule(lr){10-15}
& & & PSNR↑ & SSIM↑ & LPIPS↓ & PSNR↑ & SSIM↑ & LPIPS↓ & PSNR↑ & SSIM↑ & LPIPS↓ & PSNR↑ & SSIM↑ & LPIPS↓ \\
& & & \multicolumn{3}{c|}{Static} & \multicolumn{3}{c|}{Dynamic} & \multicolumn{3}{c|}{Static} & \multicolumn{3}{c}{Dynamic} \\
\midrule
BasicVSR++ & 9.76  & 0.14 & 23.13 & 0.651 & 0.267 & 21.31 & 0.618 & 0.402 & 27.86 & 0.920 & 0.160 & 24.09 & 0.818 & 0.236 \\
TMT        & 26.04 & 0.667 & 22.96 & 0.654 & 0.274 & 21.37 & 0.625 & 0.393 & 18.52 & 0.623 & 0.321 & 23.16 & 0.794 & 0.231 \\
DATUM      & 5.754  & 0.031 & 22.77 & 0.647 & 0.286 & 21.43 & 0.620 & 0.383 & 27.41 & 0.915 & 0.172 & 24.21 & 0.821 & 0.255 \\
DeTurb     & 58.79 & 1.24 & --    & --    & --    & --    & --    & --    & 27.84 & 0.857 & 0.175 & 22.41 & 0.797 & \underline{0.175} \\
MamabaTM     & 6.904 & 0.018 &  23.78  &   \underline{0.671}  &  0.265  & 21.44    & \underline{0.674}    & \underline{0.367}    & 28.03 & 0.841 & 0.342 & 24.42 & 0.708 & 0.295 \\
MAMAT      & 2.832  & 0.089 & \underline{23.97} & 0.663 & \underline{0.258} & \underline{21.91} & 0.623 & 0.388 & \underline{28.85} & \underline{0.937} & \underline{0.155} & \underline{24.76} & \underline{0.825} & 0.228 \\
\midrule
\textbf{RMFAT (ours)} 
           & \textbf{2.602} & \textbf{0.008} 
           & \textbf{24.01} & \textbf{0.678} & \textbf{0.251} 
           & \textbf{22.42} & \textbf{0.686} & \textbf{0.365} 
           & \textbf{28.95} & \textbf{0.941} & \textbf{0.148} 
           & \textbf{25.01} & \textbf{0.832} & \textbf{0.188} \\
\bottomrule
\end{tabular}
}
\caption{Combined performance and efficiency comparison on synthetic and real datasets (CLEAR). Bold and underlined values denote the best and second-best results, respectively. `--' indicates unavailable synthetic results for DeTurb. Speed is measured in frames per second on images with a resolution of 480$\times$384 pixels using an NVIDIA RTX 4090.}
\label{tab:comparison_combined}
\end{table*}

\subsection{Performance on synthetic datasets}

As shown in Table~\ref{tab:comparison_combined}, our RMFAT outperforms existing methods in all evaluation metrics for both static and dynamic scenes, demonstrating consistent and robust performance. In static scenes, despite limited overall improvement, RMFAT still demonstrates certain advantages in image quality and structural restoration. In dynamic scenes, due to more complex motion, RMFAT shows more significant improvements, particularly in structural similarity and perceptual consistency, reflecting its strong robustness in handling complex temporal perturbations.

In addition to its advantages in restoration quality, RMFAT also has extremely high inference efficiency. Among all the methods compared, it has the fewest parameters and the shortest single-frame inference time, indicating that the method effectively reduces computational overhead while maintaining performance, and has the potential for real-time deployment on resource-constrained devices.

Figure~\ref{fig:synthetic} shows a visual comparison of the methods on a synthetic dataset. Compared to other methods, our RMFAT can reconstruct object edges more clearly and restore more details, with subjective results closer to real images. Specifically, DATUM and MAMAT still exhibit some blurring, while BasicVSR++ and TMT perform poorly in edge restoration. The above subjective evaluation further validates RMFAT's leading advantages in structural clarity and perceptual consistency.

\subsection{Performance on real datasets}
To further validate the generalization capability of our method in real-world conditions, we evaluate it on the \textbf{CLEAR dataset}~\cite{Anantrasirichai:Atmospheric:2013}, which contains authentic atmospheric turbulence distortions captured in uncontrolled environments. This dataset come with pseudo ground truth generated using complex wavelet based image fusion; hence, objective assessment is possible. We use five static and five dynamic scenes from the dataset.

As shown in Table~\ref{tab:comparison_combined}, our RMFAT maintains leading performance across all metrics and scenarios. In static scenarios, its overall performance surpasses existing state-of-the-art methods; in dynamic scenarios, its advantages are even more pronounced, particularly in terms of perceived quality, demonstrating stronger adaptability. These results indicate that RMFAT possesses excellent generalisation capabilities and can effectively handle complex disturbances in real-world environments.

Figure~\ref{fig:real} further demonstrates visual restoration comparisons across multiple real sequences of CLEAR dataset. RMFAT can more accurately restore object details and structural edges, such as vehicle body text, human contours, and road sign content, showing significantly better visual consistency than other methods. In contrast, while DeTurb and MAMAT can retain some details, they still suffer from issues such as blurring or structural distortion. Due to RMFAT’s recurrent design, its main limitation arises when key objects appear only in a few frames. For example, in the third row (black car scene), the number plate remains blurry as the car moves rapidly toward the camera, offering limited temporal information for the recurrent process. Although DeTurb produces the sharpest result for the number plate, its model is ten times larger than RMFAT. Overall, RMFAT demonstrates stable and high-quality restoration capabilities across different content types and distortion intensities, fully showcasing its practicality and robustness in real-world applications.

Figure~\ref{fig:easytohard} shows a qualitative comparison of three real-world sequences of \textbf{ATD dataset}\footnote{\url{https://zenodo.org/records/13737763}} at different turbulence difficulty levels. Compared to existing methods, RMFAT can restore clearer structures and textures at different degradation intensities, with particularly outstanding performance in high-difficulty scenes. For example, in the most challenging bottom scene, strong turbulence causes significant geometric deformation and blurring. While most methods perform poorly in this scenario, RMFAT can still reconstruct sharper edges and more complete structural contours (such as the edges of the car and the background boundaries). In contrast, the results from TSRWGAN and DeTurb still exhibit noticeable blurring and structural distortion, while MambaTM and MAMAT, though better at preserving structure, tend to introduce artifacts or over-smoothing. These results further validate the robustness and effectiveness of RMFAT in complex real-world scenarios.

\begin{figure}[t]
    \centering
    \includegraphics[width=0.95\linewidth]{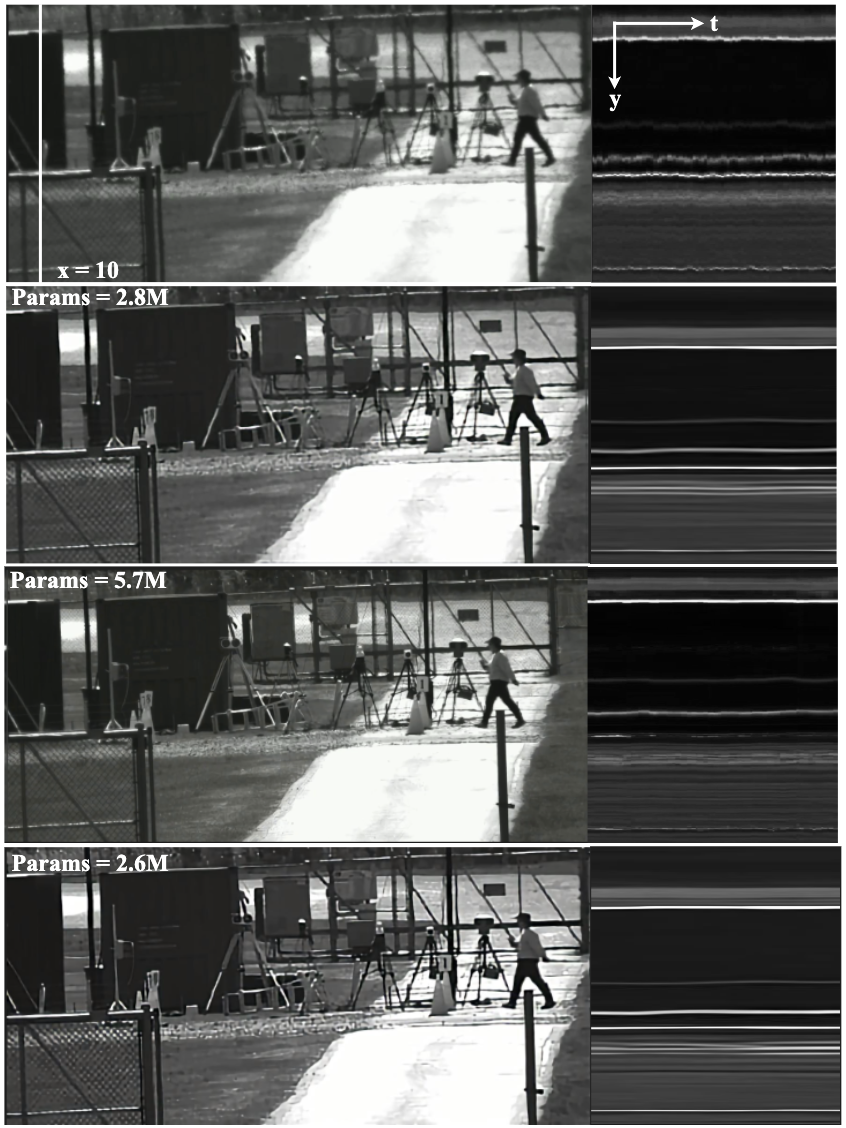}
    \caption{A real-world \textbf{BRIAR} sequence. From top to bottom: distorted input, results of the multi-frame recovery method, results of the DATUM method and results of our proposed recursive two-frame method. The left and right sides show the recovered frames and the corresponding $y$–$t$ slices at column $x = 10$, respectively.
    }
    \label{fig:feature2}
\end{figure}

We further evaluated our method on the real-world \textbf{BRIAR dataset}~\cite{Cornett_2023_WACV}, as shown in Figure~\ref{fig:feature2}. The restored image (left) and the corresponding $y$–$t$ plane (right) demonstrate that our approach achieves a strong balance between structural clarity and model simplicity, while maintaining high temporal consistency. Notably, this is accomplished using only 2.6M parameters—significantly fewer than competing methods—highlighting the efficiency of our model, especially for real-time deployment scenarios.

\subsection{Ablation study}
To assess the impact of each module in the RMFAT model on performance, we conducted ablation experiments on a synthetic dataset. Table~\ref{tab:ablation} shows the performance changes after removing different loss functions and structural components, with evaluation metrics including PSNR, SSIM, and LPIPS.

\vspace{0.5em}
\noindent\textbf{Loss function analysis.}
In terms of loss functions, removing the wavelet loss results in an increase in LPIPS from 0.365 to 0.384 (5.21\%), indicating its positive role in enhancing perceived quality. Excluding the detection loss results in a slight decrease in all metrics with SSIM decreasing from 0.686 to 0.675 (0.73\%) and LPIPS increasing slightly to 0.369 (1.10\%). This suggests that semantic supervision provides some benefit, its contribution is relatively limited, and may be omitted in lightweight configurations depending on deployment constraints. In contrast, removing flow-based consistency loss results the most significant performance drop among loss components with SSIM decreasing from 0.686 to 0.660 (3.79\%) and LPIPS rising to 0.395 (+8.22\%). These findings validate the critical role of of flow-guided supervision in preserving temporal consistency and visual coherence across frames.

\vspace{0.5em}
\noindent\textbf{Structural component analysis.}
In terms of structural modules, removing multi-scale alignment results in a significant performance decline, with PSNR dropping from 22.42 to 21.98 (1.96\%) and SSIM decreasing by 4.96\%. while LPIPS increased by 10.14\%, indicating that this module plays a crucial role in enhancing spatial alignment and feature fusion. Removing the warp operation in the decoder stage also leads to significant performance degradation, particularly in terms of perceptual quality, with LPIPS increasing by 7.12\% (from 0.365 to 0.391), indicating that feature alignment in the final stage plays a positive role in detail restoration and visual consistency. Furthermore, removing the recurrent structure has the most significant impact on performance, with SSIM dropping to 0.599 (12.71\%) and LPIPS increasing to 0.454 (24.38\%), validating the critical role of temporal modelling in maintaining video restoration consistency.

\vspace{0.5em}
\noindent The ablation study demonstrates that each component of RMFAT contributes positively to the overall restoration quality. The full model achieves the best performance across all metrics, validating the effectiveness of the proposed design.

\begin{table}[t!]
\setlength{\tabcolsep}{4.0pt}
\centering
\small
\begin{tabular}{lcccc}
\toprule
\textbf{Config} & \textbf{PSNR} ↑ & \textbf{SSIM} ↑ & \textbf{LPIPS} ↓ \\
\midrule
\multicolumn{4}{l}{\textit{Loss Component}} \\
\midrule
w/o wavelet        & 22.31 & 0.672 & 0.384 \\
w/o detection      & 22.41 & 0.681 & 0.369\\
w/o flow-based     & 22.10 & 0.660 & 0.395\\
\midrule
\multicolumn{4}{l}{\textit{Architectural}} \\
\midrule
w/o warp at decoder  & 22.14 & 0.667 & 0.391\\
w/o multi-scale warp & 21.98 & 0.652 & 0.402\\
w/o recurrent        & 20.74 & 0.599 & 0.454 \\
\midrule
\textbf{Full (Ours)} & \textbf{22.42} & \textbf{0.686} & \textbf{0.365}\\
\bottomrule
\end{tabular}
\caption{Ablation study on loss components and architecture on the dynamic real-world data.}
\label{tab:ablation}
\end{table}

\section{Conclusion}
This paper proposes a lightweight recurrent model, RMFAT, for atmospheric turbulence video restoration. By reformulating the multi-frame restoration task into a two-frame recurrent process, RMFAT significantly reduces computational costs while maintaining strong temporal modelling capabilities. The model integrates multi-scale feature extraction and alignment, and designs a loss function guided by frequency domain, consistency, and semantic information, resulting in superior recovery results in terms of spatial detail and temporal coherence.

Experiments on synthetic and real datasets demonstrate that RMFAT outperforms existing methods in terms of recovery quality and achieves over a fourfold improvement in inference speed. The model requires only two input frames and has only 2.6 million parameters, offering good real-time performance and deployment friendliness, making it suitable for scenarios with high requirements for low latency and resource constraints, such as remote sensing and surveillance.

Future work could explore alternative flow alignment mechanisms or joint multi-task training to further enhance performance and generalisation.

\bibliography{aaai2026}

\setlength{\leftmargini}{20pt}
\makeatletter\def\@listi{\leftmargin\leftmargini \topsep .5em \parsep .5em \itemsep .5em}
\def\@listii{\leftmargin\leftmarginii \labelwidth\leftmarginii \advance\labelwidth-\labelsep \topsep .4em \parsep .4em \itemsep .4em}
\def\@listiii{\leftmargin\leftmarginiii \labelwidth\leftmarginiii \advance\labelwidth-\labelsep \topsep .4em \parsep .4em \itemsep .4em}\makeatother

\setcounter{secnumdepth}{0}
\renewcommand\thesubsection{\arabic{subsection}}
\renewcommand\labelenumi{\thesubsection.\arabic{enumi}}

\newcounter{checksubsection}
\newcounter{checkitem}[checksubsection]

\newcommand{\checksubsection}[1]{%
  \refstepcounter{checksubsection}%
  \paragraph{\arabic{checksubsection}. #1}%
  \setcounter{checkitem}{0}%
}

\newcommand{\checkitem}{%
  \refstepcounter{checkitem}%
  \item[\arabic{checksubsection}.\arabic{checkitem}.]%
}
\newcommand{\question}[2]{\normalcolor\checkitem #1 #2 \color{blue}}
\newcommand{\ifyespoints}[1]{\makebox[0pt][l]{\hspace{-15pt}\normalcolor #1}}

\end{document}